\pdfoutput=1
\documentclass[11pt]{article}

\usepackage[preprint]{acl}

\usepackage{times}
\usepackage{latexsym}

\usepackage[T1]{fontenc}
\usepackage[utf8]{inputenc}

\usepackage{microtype}

\usepackage{inconsolata}

\usepackage{graphicx}

\usepackage{enumitem}

\usepackage{tikz}

\usepackage{booktabs}
\usepackage{multirow}

\usepackage{colortbl}

\definecolor{gglblue}{HTML}{4285F4}
\definecolor{gglgreen}{HTML}{0F9D58}
\definecolor{blue}{HTML}{1155cc}
\definecolor{yellow}{HTML}{e69138}
\definecolor{green}{HTML}{38761d}

\usepackage{subcaption}

\usepackage{changepage}

\usepackage{gb4e}
\noautomath

\newcommand{\namedref}[2]{\hyperref[#2]{#1~\ref*{#2}}}
\newcommand{\sectionref}[1]{\namedref{Section}{#1}}
\newcommand{\tableref}[1]{\namedref{Table}{#1}}
\newcommand{\figureref}[1]{\namedref{Figure}{#1}}
\newcommand{\appendixref}[1]{\namedref{Appendix}{#1}}

\newcommand{\eqref}[1]{\namedref{Equation}{#1}}

\newcommand{\sxs}{\textsc{s$\times$s}}

\newcommand{\psxsmqm}{MQM}
\newcommand{\sxsmqm}{\textsc{s$\times$s MQM}}
\newcommand{\sxsqr}{\textsc{s$\times$s RR}}

\newcommand{\minor}{\texttt{minor}}
\newcommand{\major}{\texttt{major}}

\newcommand{\ZhEn}{\texttt{ZhEn}}
\newcommand{\EnDe}{\texttt{EnDe}}

\title{Enhancing Human Evaluation in Machine Translation with Comparative Judgment}

\author{Yixiao Song$^\spadesuit$\thanks{ Work done during an internship at Google Translate.} \quad Parker Riley$^\diamondsuit$ \quad  Daniel Deutsch$^\diamondsuit$ \quad Markus Freitag$^\diamondsuit$\\
$^\spadesuit$Manning College of Information and Computer Sciences, UMass Amherst\\
$^\diamondsuit$Google\\
\texttt{yixiaosong@umass.edu}
}

\begin{document}
\maketitle

\begin{abstract}

Human evaluation is crucial for assessing rapidly evolving language models but is influenced by annotator proficiency and task design. This study explores the integration of comparative judgment into human annotation for machine translation (MT) and evaluates three annotation setups---point-wise Multidimensional Quality Metrics (MQM), side-by-side (\sxs) MQM, and its simplified version \sxs~relative ranking (RR). In MQM, annotators mark error spans with categories and severity levels. \sxsmqm~extends MQM to pairwise error annotation for two translations of the same input, while \sxsqr~focuses on selecting the better output without labeling errors.

Key findings are: (1) the \sxs~settings achieve higher inter-annotator agreement than MQM; (2) \sxsmqm~enhances inter-translation error marking consistency compared to MQM by, on average, 38.5\% for explicitly compared MT systems and 19.5\% for others; (3) all annotation settings return stable system rankings, with \sxsqr~offering a more efficient alternative to (\sxs) MQM;
% , albeit with higher tie rates and less granularity; 
(4) the \sxs~settings highlight subtle errors overlooked in MQM without altering absolute system evaluations.

To spur further research, we release the triply annotated datasets comprising 377 \ZhEn~and 104 \EnDe~annotation examples.\footnote{Data will be available at \url{https://github.com/google/wmt-mqm-human-evaluation/tree/main/generalMT2023}.}
% doing  annotations side-by-side does not affect the absolute evaluation of individual systems but can highlight errors that are hard to find in point-wise MQM.

\end{abstract}

\section{Introduction}\label{sec:intro}
% With the rapid advancement of more powerful language models, the quality of their generated outputs has significantly improved, often surpassing the capabilities of existing automatic evaluation metrics \citep{karpinska-iyyer-2023-large, pham2024suri}. As a result, human evaluation continues to play a vital role in assessing models' performance.
With the rapid improvement of large language models' capabilities, automatic evaluation metrics have struggled to reliably measure their quality \citep{karpinska-iyyer-2023-large, pham2024suri}. As a result, human evaluation continues to play a vital role in assessing models' performance.

Human annotations can be influenced by several difficult-to-control factors, such as annotators' proficiency and their relative leniency or stringency. Annotator proficiency can be managed by hiring experts \citep{karpinska-etal-2021-perils, krishna-etal-2023-longeval}. Varying degrees of leniency or stringency can be mitigated by carefully assigning tasks to annotators in a structured manner \citep{riley-etal-2024-finding}. 
%However, there can be other factors that affect human behavior, for example, annotation setups. 
However, other factors can affect rater behavior, such as the specific annotation task used to measure quality \cite{belz-kow-2010-comparing}.

% However, despite clear guidelines, annotators often apply their own interpretations of the standards, leading to inconsistencies and subjective variations both intra- and inter-annotators (\citealp{ACJ}; \sectionref{sec:inter-TC}).

\begin{figure*}
    \centering
    \includegraphics[scale=0.63]{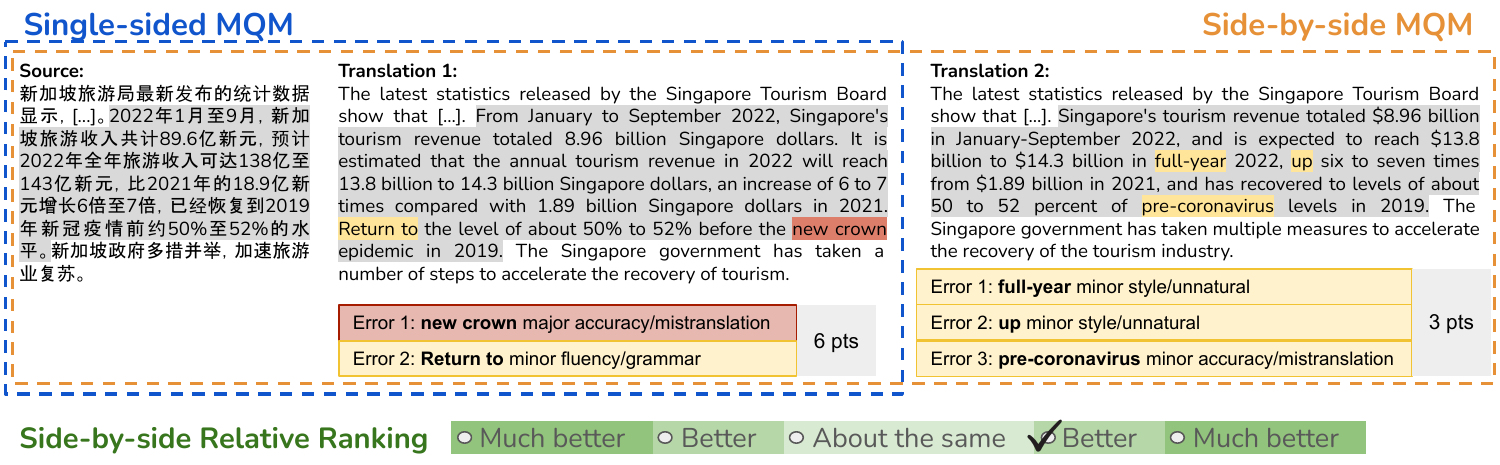}
    \caption{Illustration of the three annotation settings studied in this work (\sectionref{sec:three_studied_settings}). The grey-highlighted text
    % \prcomment{The grey highlights are not really visible when printed, even in color}
    is the segment to be annotated within their context. In {\color{blue}\textbf{single-sided}} and {\color{yellow}\textbf{side-by-side}} MQM, annotators mark error spans and assign error category with severity. The score of a segment/document is determined by the category and severity of its error(s). In {\color{green}\textbf{side-by-side relative ranking}}, annotators read two translations and choose the (much) better side or decide if they tie, without labelling errors. The scoring scheme of each setting is in \sectionref{sec:result_score_calc}.
    }
    \label{fig:figure1}
\end{figure*}

This work investigates the influence of annotation settings on annotator behavior and results by using Chinese to English (\ZhEn) and English to German (\EnDe) machine translation (MT) as a case study. It examines three annotation settings: (1) the state-of-the-art point-wise MT annotation setup \textbf{MQM} \citep{lommel2014mqm, freitag-etal-2021-experts} where annotators see one translation at a time and identify errors with category and severity assignment of each, (2) \textbf{side-by-side} (\sxs) \textbf{MQM}, where annotators see two translations of the same input at a time and give fine-grained error annotations as MQM, (3) \textbf{\sxs~relative ranking} (RR), where annotators see two translations and decide which one is better, without error annotation. The latter two settings incorporate comparative judgment \citep{LCJ1927}, a pair-wise setting that allows annotators to make relative assessments between system outputs; in \sxsqr, it assists annotators in making comparisons, while in \sxsmqm~it helps them detect errors more easily, particularly those appearing in only one output. Comparative judgment has been shown to reduce subjectivity and enhance consistency in quality judgments \citep{karpinska-etal-2021-perils, de-Moira-2022, Jones-2024}. The three settings are illustrated in \figureref{fig:figure1}.

% \sxsmqm~and \sxsqr~incorporate comparative judgment \citep{LCJ1927} \prcomment{Does SxS MQM really qualify as comparative judgment? Does that term imply a ranking/preference judgment? Is "pairwise" what we're looking for?} that allows annotators to make relative assessments between system outputs and has been shown to reduce subjectivity and enhance consistency in quality judgments \citep{karpinska-etal-2021-perils, de-Moira-2022, Jones-2024}. The three settings are illustrated in \figureref{fig:figure1}.

This work meta-evaluates human annotation results from the studied annotation settings across five aspects: inter-annotator agreement, inter-translation error marking consistency, segment- and system-level quality rankings, and error distribution in MQM and \sxsmqm. Overall, our human annotation results reveal the following relative strength/weakness of each protocol. \textbf{Comparative judgment improves inter-annotator agreement}, especially in \sxsmqm, compared to point-wise MQM. \textbf{Comparative judgment boosts inter-translation error marking consistency}, with average increases of 38.05\% for systems that were shown side-by-side, and 19.5\% for system pairs that were not. For MT system ranking, \textbf{\sxsmqm~proved more reliable in identifying equal-quality translations} while \sxsqr~offered a cost-effective alternative 
with limited ability in capturing subtle quality differences due to the lack of error annotation.\footnote{We estimate that \sxsqr~reduces costs by approximately one-third compared to MQM.}
% \prcomment{This claim is contradictory: identifying equal-quality translations is presented as good, but higher tie rates is presented as bad}, 
% limiting its ability to capture subtle differences. 
In terms of error distribution, \textbf{\sxsmqm~can affect the distribution of error categories and severities}, as seen in the higher detection rate of accuracy errors in \ZhEn~compared to MQM. Overall, the contributions of this paper are:
% \prcomment{I don't think we can claim this; let's discuss in Chat}
% In terms of error distribution, \yscomment{\ZhEn~and \EnDe~behave differently, with fluency errors more prevalent in \EnDe~and \sxsmqm~highlighting more \ZhEn~accuracy errors. The higher prevalence of accuracy errors in \ZhEn~is likely due to the linguistic differences between Chinese and English, suggesting that \textbf{\sxsmqm~tends to amplify accuracy errors in typologically distant language pairs}.}
% \textbf{\sxsmqm~highlighted more major accuracy errors in \ZhEn}, 
% \prcomment{Should this be reframed as something like "error distributions were different between (pointwise) MQM and SxS MQM"?},
% likely due to linguistic differences between Chinese and English, while \EnDe~annotations showed a higher prevalence of fluency errors. 

\vspace{-3pt}
\begin{enumerate}[label=(\arabic*), itemsep=-0.2em]
    \item It examines the  point-wise MQM, \sxsmqm, and the simplified \sxsqr, offering a systematic investigation of comparative judgment in human annotation tasks in MT;
    
    \item It offers insights that \sxsmqm~provides more reliable and fine-grained annotations while \sxsqr~provides an efficient alternative when detailed error annotation is not required.
\end{enumerate}
\vspace{-6pt}

% The human evaluation data and code will be open-sourced for future research.% \footnote{\url{https://github.com/google-research/google-research/tree/master/anthea}}

\section{Annotation settings}\label{sec:annotation_settings}

This section introduces the core concepts and methodologies used in this study. It begins with key terminology in \sectionref{sec:term}, followed by the rationale for integrating comparative judgment into annotation tasks (\sectionref{sec:comparative_judgement}) and an overview of the MQM framework in \sectionref{sec:MQM}. The three annotation settings analyzed in this work--—MQM, \sxsmqm, and \sxsqr—--are then described in detail in \sectionref{sec:three_studied_settings}. Finally, \sectionref{sec:result_score_calc} explains how annotations are converted into numeric scores.

\subsection{Terminology}\label{sec:term}

Two terms in the current work are defined here following \citet{riley-etal-2024-finding}.  A \textbf{segment} is a unit of one or multiple sentences that is highlighted at a time for annotators to focus on for annotation (i.e., the grey-highlighted text in \figureref{fig:figure1}); a \textbf{document} is a sequence of input segments (e.g., an excerpt from an article).

% \begin{itemize}[label=-, left=0pt, itemsep=0pt]
%     \item \textbf{Segment} (seg): a unit of one or multiple sentences that is highlighted at a time for annotators to focus on for annotation;
    
%     \item \textbf{Document} (doc): A sequence of input segments (e.g., an excerpt from an article).
% \end{itemize}

\subsection{Side-by-side annotation}\label{sec:comparative_judgement}

Comparative judgment \citep{LCJ1927} is a psychometric method that presents two items side by side, asking which better satisfies a given criterion. It assumes that people are more reliable when comparing two items than evaluating them individually. Studies show that comparative judgment improves the consistency and accuracy of teachers' assessment of students' writing \citep{ACJ, de-Moira-2022, Jones-2024}. 

In natural language generation, the \sxs~annotation (more commonly called \textit{pair-wise annotation}) has been used for tasks like open-ended text generation \citep{wang-etal-2023-knn, krishna-etal-2023-longeval}, machine translation \citep{karpinska-iyyer-2023-large}, and long-form question answering \citep{xu-etal-2023-critical}. However, there lacks systematic study that compares this approach with point-wise approaches in terms of annotator behavior and annotation results.

\subsection{Multidimensional quality metrics}\label{sec:MQM}

MQM is an annotation framework proposed by \citet{lommel2014mqm} and refined by \citet{freitag-etal-2021-experts}. It is the state-of-the-art annotation setting currently used by WMT \citep{freitag-etal-2023-results}. It involves marking error spans and assigning error severity and category. In this work, annotators use the error hierarchy in \tableref{tab:error_hierarchy} (\appendixref{appendix:error_hierarchy}) for error categorization. All categories can be either \major~or \minor, except for \texttt{Non-translation} which is always \major.\footnote{\texttt{Major} errors significantly alter the meaning of the source text; \minor~errors are noticeable but do not significantly alter the source meaning.} Annotators are instructed that the more precise the error spans, the more informative the annotation.

\subsection{Studied annotation settings}\label{sec:three_studied_settings}

Three annotation settings, illustrated in \figureref{fig:figure1}, are studied and compared to each other: \psxsmqm, \sxsmqm, and \sxs~RR. Annotators evaluate system outputs segment by segment, with full access to the surrounding context.

\textbf{Single-sided MQM} (i.e., point-wise MQM; henceforth, MQM), as in \sectionref{sec:MQM}, involves annotators evaluating one translation at a time. 

% is the setting described in Section~\ref{sec:MQM}. Each annotator reviews a single translation at a time.%uses the within-subject setting proposed by \citet{riley-etal-2024-finding} where all system outputs of a given input are annotated by the same annotator(s). Although annotators see one translation of a document at a time, it controls the noise from annotators as the same degree of leniency/stringency is applied to all system outputs of a given input. \citet{riley-etal-2024-finding} shows that this within-subject \psxsmqm~yields high stability of system ranking.

\textbf{\sxsmqm}~is the same as MQM with one difference: two translations (each from a different system) are shown side-by-side instead of just one. The core annotation task is otherwise unchanged.
%builds upon the within-subject \psxsmqm~with the modification that two translations from two systems are shown side-by-side to annotators, along with the input. The core annotation task is the same as \psxsmqm.
% In principle, \sxsmqm~should enhance annotators' consistency of error span and category marking, compared to MQM, by enabling direct comparisons between the two systems. % No matter what interpretation an annotator has of the guidelines at the time of annotation, the interpretation is applied to both system outputs in the same way, unlike \psxsmqm~that still leaves room for variation. 

% \textbf{\sxsqr} uses the within-subject setup without requiring error marking and categorization. 

\textbf{\sxsqr} does not require detailed error annotation. Annotators view two translations of the same input text side-by-side and rate them on a five-point scale as \texttt{(much) better} on one side or \texttt{about the same} (see \figureref{fig:figure1}). This setting evaluates whether side-by-side annotation can provide reliable system rankings without detailed error annotation.

% \vspace{6pt}

% \begin{center}
% \begin{tikzpicture}[scale=0.75]

%     % Define the scale points
%     \foreach \i/\label in {0/1, 2/2, 4/3, 6/4, 8/5} {
%         \node[circle, draw, inner sep=2pt] at (\i,0) {};
%     }

%     % Define the labels below the points
%     \node[scale=0.7] at (0,-0.8) {\texttt{Much better}};
%     \node[scale=0.7] at (2,-0.8) {\texttt{Better}};
%     \node[scale=0.7] at (4,-0.8) {\texttt{About the same}};
%     \node[scale=0.7] at (6,-0.8) {\texttt{Better}};
%     \node[scale=0.7] at (8,-0.8) {\texttt{Much better}};

%     % Draw lines between the points
%     \draw (0.15,0) -- (1.85,0);
%     \draw (2.15,0) -- (3.85,0);
%     \draw (4.15,0) -- (5.85,0);
%     \draw (6.15,0) -- (7.85,0);

% \end{tikzpicture}
% \end{center}

\subsection{Score calculation}\label{sec:result_score_calc}

Expert annotations are converted into numeric scores for system and annotation setting comparison. For MQM and \sxsmqm, the segment score is the average of the scores assigned by each annotator, with each score determined by the error severity and category (\tableref{tab:weighting-scheme} in \appendixref{appendix:scoring_scheme}). The system score is obtained by averaging the segment scores. For MQM, lower scores are better, with an error-free segment receiving a score of 0. The scores are \textit{z}-normalized following \citet{riley-etal-2024-finding}.

The scoring of \sxsqr~follows MQM in that a lower score is better. In each rated pair, the \texttt{much worse} translation segment is penalized by 2 points, and the \texttt{worse} segment by 1 point. If both translations are of similar quality, no penalty is applied. As the MQM settings, the segment score is the average of the scores assigned by each annotator. The system score is averaged over the segment scores.\footnote{We z-normalized the (\sxs) MQM scores to account for variations in score ranges across different annotators, ensuring comparability and mitigating individual annotator biases. In contrast, the \sxsqr~scores were not z-normalized since they are inherently constrained within the range of -2 to 0.}

\section{Experimental setup}\label{sec:exp_setup}

This section details the setup of human experiments, including the dataset, the criteria for selecting systems to be evaluated, and the process for assigning tasks among annotators.

\subsection{Dataset and language pairs}

The human annotation experiments are performed on the system outputs from the news domain in WMT2023 \citep{freitag-etal-2023-results}, covering two language pairs: Chinese to English (\ZhEn) and English to German (\EnDe). Basic statistics are in \tableref{tab:basic_stats}.

\begin{table}[ht]
\fontsize{6}{7}\selectfont
\centering
\resizebox{0.75\columnwidth}{!}{%
\begin{tabular}{@{}p{2.1cm}p{0.5cm}p{0.5cm}@{}}
\midrule
               & \ZhEn & \EnDe \\ \cmidrule{1-3}
Documents      & 38            & 30 \\
Segments      & 377           & 104 \\
Avg. English tokens/seg   &     32.02     & 71.91\\\midrule
\end{tabular}%
}
\caption{
Basic dataset statistics. For \ZhEn, average tokens per segment are based on the English reference translation, and for \EnDe, on the English source. Tokens are counted using whitespace in both cases.
}
\label{tab:basic_stats}
\vspace{-10pt}
\end{table}

\subsection{System pairs}\label{sec:sys_pairs}

Due to the time and cost involved in human evaluation, pairwise comparisons of all MT systems in the \sxs~settings are impractical. Therefore, 5 system pairs are selected per language pair (\tableref{tab:sys_pairs}), which are drawn from the systems in the WMT 2023 General Machine Translation Task \citep{kocmi-etal-2023-findings}.%, with a focus on systems with similar quality, as distinguishing systems with smaller quality differences presents a greater challenge in practical scenarios. 
% , focusing on systems with similar quality, as distinguishing between systems with smaller quality differences presents a greater challenge in practical scenarios.

\begin{table}[ht]
\fontsize{7}{8}\selectfont
\centering
\resizebox{0.92\columnwidth}{!}{%
\begin{tabular}{@{}p{0.1cm}p{1.51cm}cp{0.3cm}p{0.4cm}p{2.1cm}@{}}
\midrule
        & \textbf{System Pairs} & \textbf{Rank} & \textbf{\textit{p}} &  \textbf{Cross-BLEU} & \textbf{Criteria} \\ \midrule
\multirow{10}{*}{\rotatebox{90}{\ZhEn}} & GPT4-5shot &  1  & \multirow{2}{*}{0.05} & \multirow{2}{*}{62.2} & \multirow{2}{*}{Top 2 systems} \\
                                       & Lan-BridgeMT & 2 & & & \\\cmidrule{2-6}
                                       & HW-TSC & 4 & \multirow{2}{*}{0.27} & \multirow{2}{*}{57.3} & \multirow{4}{*}{High text similarity} \\
                                       & ONLINE-A & 5 & & & \\\cmidrule{2-5}
                                       & IOL-Research & 6  & \multirow{2}{*}{0.17} & \multirow{2}{*}{52.2} & \\
                                       & ONLINE-B & 8 & & & \\\cmidrule{2-6}
                                       & ONLINE-W & 10 & \multirow{2}{*}{0.10} & \multirow{2}{*}{31.2} & \multirow{4}{*}{Lower text similarity} \\
                                       & NLLB\_Greedy & 12 & & & \\\cmidrule{2-5}
                                       & NLLB\_BLEU & 14 & \multirow{2}{*}{0.40} & \multirow{2}{*}{35.7} & \\
                                       & ONLINE-M & 15 &  & & \\\midrule
\multirow{10}{*}{\rotatebox{90}{\EnDe}} & ONLINE-W  & 1   & \multirow{2}{*}{0.09} & \multirow{2}{*}{53.1} & \multirow{2}{*}{Top 2 Systems} \\  
                                       & GPT4-5shot  & 2  & & & \\\cmidrule{2-6}
                                       & ONLINE-Y   & 5   & \multirow{2}{*}{0.48} & \multirow{2}{*}{62.0} &  \multirow{4}{*}{High text similarity}\\
                                       & ONLINE-A   & 6   & & & \\\cmidrule{2-5}
                                       & ONLINE-M   & 7   & \multirow{2}{*}{0.22} & \multirow{2}{*}{56.3} & \\
                                       & ONLINE-G   & 8   & & & \\\cmidrule{2-6}
                                       & GPT4-5shot & 2  & \multirow{2}{*}{0.11} & \multirow{2}{*}{39.3} & \multirow{4}{*}{Lower text similarity} \\
                                       & refA       & 3   & & & \\\cmidrule{2-5}
                                       & NLLB\_BLEU & 10 & \multirow{2}{*}{0.24} & \multirow{2}{*}{44.3} & \\
                                       & Lan-BridgeMT  & 11 & & & \\\midrule
\end{tabular}%
}
\caption{Systems annotated in the human experiments. The ranks are determined by XCOMET. The $p$ values are calculated by a random permutation test with 10000 trials to determine quality similarity. Cross-BLEU quantifies the text similarity of system outputs.
}
\label{tab:sys_pairs}
\vspace{-3pt}
\end{table}

Two features are considered when forming system pairs: text similarity and quality similarity. For text similarity, cross-BLEU \citep{papineni-etal-2002-bleu} is applied. For quality similarity, XCOMET-QE-Ensemble (XCOMET) \citep{xcomet, freitag-etal-2023-results} is used in tandem with a random permutation test.\footnote{\texttt{permutation\_test} from \texttt{scipy.stats}.} XCOMET provides segment scores for each system. Two systems are considered similar in quality if the permutation test of their segment scores returns $p > 0.05$. 
% \url{https://docs.scipy.org/doc/scipy-1.14.1/reference/generated/scipy.stats.permutation_test.html}

For each language pair, the top two systems identified by XCOMET form a pair; two pairs have high text similarity (cross-BLEU); and two have low text similarity. In all cases, systems with similar quality were selected, as distinguishing small quality differences presents a greater challenge in practical scenarios. The system pairs are listed in \tableref{tab:sys_pairs}.\footnote{For brevity, NLLB\_MBR\_BLEU is referred to as NLLB\_BLEU in this work.} Applying these criteria resulted in GPT4-5shot appearing twice in \EnDe. This has the benefit of allowing comparison between each instance of this system's evaluation in the side-by-side settings.%~to examine whether a system's MQM score varies with different comparison systems.

\subsection{Task assignment}\label{sec:task_assignment}

The annotation experiments are conducted by professional translators who regularly perform MQM annotation. Tasks are distributed approximately evenly among 8 annotators for \ZhEn~ and 10 for \EnDe. To mitigate rater bias, we use the within-subject setup of \citet{riley-etal-2024-finding}: for each input document, all system translations for that document are evaluated by the same set of 3 annotators. Additionally, each translation is evaluated by the same 3 annotators in all 3 annotation settings.

\section{Meta evaluation of human evaluations}\label{sec:meta_eval_metrics}

This section outlines five criteria for analyzing the human annotation results: (1) inter-annotator agreement, (2) inter-translation consistency, (3) system-level ranking, (4) segment-level ranking, and (5) error distribution. Results are in \sectionref{sec:results_and_discussion}.

\noindent\textbf{Inter-annotator agreement} (IAA) Each segment pair is annotated by three annotators, allowing for calculating the IAA. For each segment translated by system \textit{a} and \textit{b}, each annotator's annotation is categorized as \textit{a} $>$ \textit{b}, \textit{a} $=$ \textit{b}, or \textit{a} $<$ \textit{b}.
% \footnote{The conversion is based on the segment MQM scores by each annotator for \psxsmqm~and \sxsmqm. For \sxsqr, the conversion is based on the point on the scale. \texttt{Much better} and \texttt{better} are collapsed to \texttt{better}; \texttt{Much worse} and \texttt{worse} are collapsed to \texttt{worse}.} 
Krippendorff's $\alpha$ \citep{krippendorff2018content} quantifies the IAA.\footnote{In the MQM settings, two segments tie if they have the same score.}

\noindent\textbf{Inter-translation consistency} When the same error occurs in translations from multiple systems of the same source input, annotators should mark it consistently with the same span, category, and severity. This consistency is crucial for fair system comparisons and training MQM-style automatic metrics \citep{juraska-etal-2023-metricx, fernandes-etal-2023-devil}. Inter-translation consistency quantifies the degree to which annotators achieve this uniformity across translations. The detailed process for calculating this consistency is provided in \appendixref{appendix:itc_pseudo_code}.

\noindent\textbf{Agreement in segment-level rankings} Pairwise ranking agreement (PRA) \citep{deutsch-etal-2023-ties}, defined in \eqref{eq:PRA} and \tableref{tab:pra_terms} (\appendixref{appendix:pra_terms}), measures the consistency between two annotation settings in ranking translation pairs by considering agreements, disagreements, and ties to evaluate alignment between evaluation methods.\footnote{The metric is termed \textit{pairwise accuracy} in \citet{deutsch-etal-2023-ties}. However, since there is no gold reference in this work, the metric is referred to as \textit{agreement}. The word \textit{ranking} is to emphasize that the metric pertains to rankings.}

% \prcomment{Without defining the terms, the equation is less useful. Consider moving the equation to the Appendix and replacing it with a prose description, such as ``the proportion of segment translation pairs where two metrics agree on which is better (or that they are tied).''}

% \vspace{-5pt}
% \begin{equation}
%   \resizebox{0.6\hsize}{!}{$
%   PRA = \frac{C + T_{\alpha\beta}}{C + D + T_\alpha + T_\beta + T_{\alpha\beta}}
%   $}
%   \label{eq:PRA}
% \end{equation}
% \vspace{-8pt}

\noindent\textbf{Agreement in system-level rankings} Systems in each pair in \tableref{tab:sys_pairs} are ranked pairwise based on scores calculated as detailed in \sectionref{sec:result_score_calc}. A random permutation test is applied to the segment-level scores of the paired systems to evaluate the statistical significance of observed differences.

\noindent\textbf{Error distribution} For the two MQM settings, one possibility is that they identify more/fewer errors of particular kinds. To explore this, the total number of target-side errors annotated across all 3 ratings was counted for each language pair and MQM setting by category and severity.%\footnote{It is possible that different annotators annotate the same error in a segment. For the error distribution calculation, such errors are not deduplicated.}

% The next section presents the results of the meta evaluation of the human annotations.

\section{Results and discussion}\label{sec:results_and_discussion}

This section presents the meta-evaluation of the human experiments using the metrics outlined in \sectionref{sec:meta_eval_metrics}, with results reported based on \textit{z}-normalized scores following \citet{riley-etal-2024-finding}. Segments annotated by a \ZhEn~outlier annotator are excluded.\footnote{The rationale for excluding the outlier annotator is provided in \appendixref{appendix:outlier_annotator}. After excluding the segments annotated by the outlier annotator, \ZhEn~has 16 documents with 220 segments and an average token count of 31.48 per segment.} The findings show that comparative judgment improves annotator agreement and consistency, maintains a reliable quality ranking, and facilitates accuracy error finding in \ZhEn.

\subsection{Inter-annotator agreement}\label{sec:annotator_agreement}

\textbf{\sxs~settings consistently yield higher agreement, particularly in \sxsmqm, in both \ZhEn~and \EnDe.} \tableref{tab:inter-AA} presents the Krippendorff's $\alpha$, indicating fair agreement among annotators in the three annotation protocols. The IAA does not exhibit a clear correlation with textual similarity between systems, as detailed in  \tableref{tab:inter-AA_expanded} (\appendixref{appendix:IAA_expanded}), an expanded version of \tableref{tab:inter-AA}.

\begin{table}[htbp]
    \fontsize{5}{6}\selectfont
    \centering
    \resizebox{0.9\columnwidth}{!}{%
    \begin{tabular}{@{}rrrr@{}}
    \midrule
     & \textbf{\psxsmqm} & \textbf{\sxsmqm} $\uparrow$ & \textbf{\sxsqr} $\uparrow$ \\ \cmidrule{1-4}
    \ZhEn & 0.2178 & \textbf{0.2510} & 0.2380 \\
    \EnDe & 0.2345 & \textbf{0.3594} & 0.2402 \\\midrule
    \end{tabular}%
    }
    \caption{Krippendorff's $\alpha$ in three annotation settings. The annotators in each setting achieve a fair agreement.
    }
    \label{tab:inter-AA}
    \vspace{-5pt}
\end{table}

The results suggest that comparative judgment improves alignment among human annotators in evaluations. This is likely because MQM, as a pointwise approach, introduces more noise by preventing direct comparisons between translations. In contrast, the \sxs~settings allow for direct comparisons, reducing noise by minimizing inconsistent error marking (\sectionref{sec:itc}) and instances where shared mistakes are flagged for one system but overlooked for the other.

We hypothesize two reasons for \sxsmqm's higher agreement compared to \sxsqr. 

First, \sxsmqm~enables explicit error marking, reducing ambiguity and enhances the clarity of the decision-making process. In contrast, \sxsqr~requires annotators to simultaneously evaluate and weigh multiple aspects of two segments (e.g., accuracy and style). This increases cognitive load and introduces greater variability in their decisions.

Second, the increased cognitive load may cause annotators to be influenced by longer segments during comparative judgments. To test this, we ranked the segments by length, divided them into three equally sized groups, and computed Krippendorff's $\alpha$ for each group. The results in \tableref{tab:inter-AA-3-buckets} (\appendixref{appendix:IAA_expanded}) show that, in the \sxs~settings, the shortest segments achieve the highest agreement.

Overall, the improvement in segment-level agreement introduced by comparative judgment is valuable because segment-level evaluation is susceptible to noise \citep{freitag-etal-2023-results}, so mitigating that noise can improve reliability.
%Overall, the improvement introduced by comparative judgment is valuable for two key factors. First, the agreement is calculated at the segment level, a method susceptible to noise due to the inherent variability of individual segments \citep{freitag-etal-2023-results}. Second, our expert annotators consistently perform MQM ratings, ensuring high-quality annotations and a strong baseline agreement, leaving limited room for further enhancement. Given these constraints, the improvement gained from comparative judgement is notable.

\subsection{Inter-translation consistency}\label{sec:itc}

\begin{table*}[ht]
\fontsize{8}{10}\selectfont
\centering
% \renewcommand{\arraystretch}{0.8} % Reduce vertical space
% \resizebox{0.7\columnwidth}{!}{%
% \begin{tabular}{crrrrr}
\begin{tabular}{p{0.5cm}p{1.8cm}p{1.8cm}p{1.8cm}p{1.8cm}p{2.5cm}}
\midrule
 & \textbf{Setting} & \textbf{Span} $\uparrow$ & \textbf{Span + Cat.} $\uparrow$ & \textbf{Span + Sev.} $\uparrow$ & \textbf{Span + Cat. + Sev.} $\uparrow$ \\ \midrule
\multicolumn{6}{c}{Inter-translation consistency from explicitly compared systems (5 pairs)} \\\midrule

\multirow{2}{*}{\ZhEn} & \psxsmqm & 26.78\% & 24.66\% & 26.18\% & 24.25\% \\
 & \cellcolor{green!20}\sxsmqm  & \cellcolor{green!20}67.50\%  & \cellcolor{green!20}67.03\%  & \cellcolor{green!20}67.26\% & \cellcolor{green!20}66.85\% \\\cmidrule{1-6}

\multirow{2}{*}{\EnDe} & \psxsmqm & 45.07\% & 41.61\% & 42.81\% & 40.25\% \\
 & \cellcolor{green!20}\sxsmqm  & \cellcolor{green!20}78.7\%  & \cellcolor{green!20}77.84\% & \cellcolor{green!20}78.48\% & \cellcolor{green!20}77.67\% \\\midrule

\multicolumn{6}{c}{Inter-translation consistency from \underline{\textit{not}} explicitly compared systems (\ZhEn: 40 pairs; \EnDe: 31 pairs)} \\\midrule
\multirow{2}{*}{\ZhEn} & \psxsmqm & 26.77\% & 24.63\% & 26.06\% & 24.15\% \\
 & \cellcolor{green!20}\sxsmqm  & \cellcolor{green!20}46.90\%  & \cellcolor{green!20}45.77\%  & \cellcolor{green!20}46.00\% & \cellcolor{green!20}45.12\% \\\cmidrule{1-6}

\multirow{2}{*}{\EnDe} & \psxsmqm & 45.60\% & 42.87\% & 43.81\% & 41.60\% \\
 & \cellcolor{green!20}\sxsmqm  & \cellcolor{green!20}63.53\%  & \cellcolor{green!20}62.39\% & \cellcolor{green!20}61.67\% & \cellcolor{green!20}60.69\% \\\midrule
\end{tabular}%
% }
\caption{Inter-translation consistency, averaged over 7 (\ZhEn) and 10 (\EnDe) annotators, in \psxsmqm~and \sxsmqm.  Cat. = category, Sev. = severity. Inter-translation consistency is calculated for four criteria of what counts as common errors in two systems, for example, Span + Cat. = errors with the same span \textit{and} category. For \EnDe, the annotation of GPT4-5shot in pair with ONLINE-W is not included in the calculation of the lower table results. The green color highlights the higher values between MQM and \sxsmqm.
}
\label{tab:inter-TC_psxs_sxs_removed}
\vspace{-10pt}
\end{table*}

\textbf{\sxsmqm~demonstrates remarkable increases in inter-translation error marking consistency}, as shown in \tableref{tab:inter-TC_psxs_sxs_removed}.\footnote{The results without removing the \ZhEn~outlier annotator are in \tableref{tab:inter-TC_psxs_sxs}, which remains similar to \tableref{tab:inter-TC_psxs_sxs_removed}, meaning that, while the outlier annotator identified significantly more errors, they also exhibited improved inter-translation consistency.}
% \prcomment{This needs to be reworded; "albeit significantly more annotated errors" is missing something to parse properly}} 
% \prcomment{Should we just use Table 14 instead of 4? More generally, does it make sense to just remove the outlier rater \textbf{everywhere}?}
The upper half of \tableref{tab:inter-TC_psxs_sxs_removed} shows a substantial consistency increase, averaging $40.4\%$ for \ZhEn~and $35.7\%$ for \EnDe, when evaluating two systems together in \sxsmqm~(i.e., the pairs in \tableref{tab:sys_pairs}). This improvement persists, averaging $20.4\%$ for \ZhEn~and $18.6\%$ for \EnDe, even when two systems are not evaluated side-by-side, as shown in the lower table.\footnote{The \ZhEn~results are the average of 40 system pairs. For \EnDe, because GPT4-5shot is annotated twice in \sxsmqm, one of the annotated GPT4-5shot is excluded. Hence, the last two rows in \tableref{tab:inter-TC_psxs_sxs} are the averages of 31 pairs.} All increases extends beyond error spans into categories and severity.
% \prcomment{The improvement for non-compared systems is interesting; can we hypothesize why it is the case?}
The improvement in non-compared systems in \sxsmqm~may result from exposure to side-by-side comparisons, which potentially refine annotators' internal error detection standards as well as increase error awareness and cognitive anchoring.

The findings demonstrate that comparative judgment significantly enhances annotators' consistency in identifying error spans and assigning error severity and categories. The improvement is valuable for both gaining insights from annotations and training MQM-style automated metrics.

\subsection{Segment-level ranking agreement}\label{sec:seg_pra}

\textbf{\sxsmqm~and \sxsqr~show solid agreement with each other and are better at identifying equal-quality segments.} \tableref{tab:segment_PRA} presents the PRA results for every pair of settings in \ZhEn~and \EnDe. \tableref{tab:tie_rate} reports the tie rates for each pair of settings. The results provide three important insights. 

\begin{table}[t]
    \fontsize{5}{6}\selectfont
    \centering
    \resizebox{0.96\columnwidth}{!}{%
    \begin{tabular}{@{}rrrrr@{}}
    \midrule
    \textbf{$\alpha$ setting} & \textbf{$\beta$ setting} & \textbf{\ZhEn~PRA} $\uparrow$ & \textbf{\EnDe~PRA} $\uparrow$ & \textbf{Avg.}\\ \cmidrule{1-5}
    \psxsmqm          & \sxsqr           & 0.568   &  0.540 & 0.554 \\
    \sxsmqm           & \sxsqr           & \textbf{0.626}   &  0.629 & 0.628 \\
    \psxsmqm          & \sxsmqm          & 0.623   &  \textbf{0.646}  & 0.635 \\\midrule
    \end{tabular}%
    }
    \caption{Segment pairwise ranking agreement between every two annotation settings. Results are based on \textit{z}-scores with the \ZhEn~outlier annotator being excluded.
    }
    \label{tab:segment_PRA}
    \vspace{-5pt}
% https://docs.google.com/spreadsheets/d/1L32sSpspzRf28XHxoN3NJPl-LrAIM-DOk-B79d1jIyU/edit?usp=sharing
% /data/yixiao/human_eval_proj/code_pra/pra.py
\end{table}

First, MQM and \sxsqr~have the lowest agreement in both language pairs, largely due to the fundamental differences in their features: point-wise \textit{vs}.\ pairwise and detailed error annotation \textit{vs}.\ preference only. This shows that methodological divergence indeed impacts annotation outcomes.

Second, \sxsmqm~and \sxsqr~show solid agreement (\tableref{tab:segment_PRA}). With better IAA than MQM (\tableref{tab:inter-AA}) and lower cost than \sxsmqm, \sxsqr~is an appealing and efficient choice when detailed error annotation is not required.

Third, the MQM setting has the lowest tie rate (\tableref{tab:tie_rate}). This can be attributed to the fact that MQM lacks explicit comparisons between paired segments, which results in its low inter-translation consistency (\tableref{tab:inter-TC_psxs_sxs_removed}). As a result, MQM may misjudge segment pairs of equal quality, compromising the reliability of its annotation outcomes.

\begin{table}[htbp]
    \fontsize{5}{6}\selectfont
    \centering
    \resizebox{0.9\columnwidth}{!}{%
    \begin{tabular}{@{}rrrr@{}}
    \midrule
    \textbf{Language pair} & \textbf{MQM} & \textbf{\sxsmqm} & \textbf{\sxsqr} \\\cmidrule{1-4}
    \ZhEn & 7.36\% & 16.55\% & 18.55\% \\
    \EnDe & 6.92\% & 11.54\% & 16.15\% \\\midrule
    \end{tabular}%
    }
    \caption{
    Tie rate in three annotation settings. \sxsqr~has the highest tie rate in both language pairs.
    }
    \label{tab:tie_rate}
    % /data/yixiao/human_eval_proj/code_pra/pra.py
\vspace{-5pt}
\end{table}

% \yscomment{psxs mqm has the lowest tie rate, due to table 6. when two seg are actually tied, psxs is less likely to find out than sxs and qr do. <-- due to how mqm is designed}

\subsection{System-level ranking agreement}\label{sec:sys_level_agreement}

\begin{table*}[t]
\centering
\begin{minipage}[t]{0.47\textwidth}
    \fontsize{6}{7}\selectfont
    \centering
    \resizebox{\columnwidth}{!}{%
    \begin{tabular}{@{}crrrr@{}}
    \midrule
     & \textbf{Setting}   & \textbf{Better System} & \textbf{Worse System} & \textbf{\textit{p} value} \\ \midrule

    \multirow{3}{*}{\rotatebox{90}{Top 2}} & \psxsmqm & Lan-BridgeMT (-0.33) & GPT4-5shot (-0.28)   & \textbf{0.013} \\
     & \sxsmqm  & Lan-BridgeMT (-0.26) & GPT4-5shot (-0.21)   & \textbf{0.025} \\
     & \sxsqr   & Lan-BridgeMT (0.36)  & GPT4-5shot (0.47)    & \textbf{0.013} \\\midrule

    \multirow{6}{*}{\rotatebox{90}{High text sim}} & \psxsmqm & HW-TSC (-0.20)       & ONLINE-A (-0.18)     & 0.277 \\
     & \sxsmqm  & HW-TSC (-0.17)       & ONLINE-A (-0.14)     & 0.234 \\
     & \sxsqr   & HW-TSC (0.33)        & ONLINE-A (0.52)      & \textbf{0.000} \\\cmidrule{2-5}

     & \psxsmqm & ONLINE-B (-0.13)     & IOL\_Research (-0.11) & 0.213 \\
     & \sxsmqm  & ONLINE-B (-0.17)     & IOL\_Research (-0.10) & \textbf{0.014} \\
     & \sxsqr   & ONLINE-B (0.35)      & IOL\_Research (0.54)  & \textbf{0.000} \\\midrule

    \multirow{6}{*}{\rotatebox{90}{Low text sim}} & \psxsmqm & ONLINE-W (0.02)     & NLLB\_Greedy (0.48) & \textbf{0.000}  \\
     & \sxsmqm  & ONLINE-W (0.02)     & NLLB\_Greedy (0.41) & \textbf{0.000}  \\
     & \sxsqr   & ONLINE-W (0.29)     & NLLB\_Greedy (0.75) & \textbf{0.000}  \\\cmidrule{2-5}

     & \psxsmqm & ONLINE-M (0.20)     & NLLB\_BLEU (0.50)   & \textbf{0.000}  \\
     & \sxsmqm  & ONLINE-M (0.19)     & NLLB\_BLEU (0.40)   & \textbf{0.000}  \\
     & \sxsqr   & ONLINE-M (0.31)     & NLLB\_BLEU (0.61)   & \textbf{0.000}  \\\midrule
    
    \end{tabular}%
    }
    \vspace{-4pt}
    \caption*{(a) Chinese $\rightarrow$ English} % Label for the left table
\end{minipage}
\hspace{0.03\textwidth} % Add space between the tables
\begin{minipage}[t]{0.47\textwidth}
    \fontsize{6}{7}\selectfont
    \centering
    \resizebox{\columnwidth}{!}{%
    \begin{tabular}{@{}crrrr@{}}
    \midrule
     & \textbf{Setting}   & \textbf{Better System} & \textbf{Worse System} & \textbf{\textit{p} value} \\ \midrule

    \multirow{3}{*}{\rotatebox{90}{Top 2}} & \psxsmqm & ONLINE-W (-0.32) & GPT4-5shot (-0.27)  &  0.075 \\
     & \sxsmqm  & ONLINE-W (-0.35) & GPT4-5shot (-0.29)  &  0.070 \\
     & \sxsqr   & ONLINE-W (0.33) & GPT4-5shot (0.45)    &  \textbf{0.046} \\\midrule

    \multirow{6}{*}{\rotatebox{90}{High text sim}} &  \psxsmqm & ONLINE-A (-0.16) & ONLINE-Y (-0.08)    &  \textbf{0.029} \\
     & \sxsmqm  & ONLINE-A (-0.18) & ONLINE-Y (-0.10)    &  \textbf{0.014} \\
     & \cellcolor{red!8}\sxsqr   & ONLINE-Y (0.38) & ONLINE-A (0.43)    &  0.258 \\\cmidrule{2-5}

     & \psxsmqm & ONLINE-M (0.02) & ONLINE-G (0.13)    &  0.058 \\
     & \sxsmqm  & ONLINE-M (0.08) & ONLINE-G (0.16)    &  0.15  \\
     & \sxsqr   & ONLINE-M (0.43) & ONLINE-G (0.44)    &  0.448 \\\midrule

    \multirow{6}{*}{\rotatebox{90}{Low text sim}} & \psxsmqm & refA (-0.35)     & GPT4-5shot (-0.27)  & \textbf{0.037}  \\
     & \sxsmqm  & refA (-0.32)     & GPT4-5shot (-0.31)  & 0.412  \\
     & \sxsqr   & refA (0.36)      & GPT4-5shot (0.50)   & \textbf{0.027}  \\\cmidrule{2-5}

     & \psxsmqm & Lan-BridgeMT (0.39) & NLLB\_BLEU (0.63)  & \textbf{0.005} \\
     & \sxsmqm  & Lan-BridgeMT (0.44) & NLLB\_BLEU (0.87)  & \textbf{0.000} \\
     & \sxsqr   & Lan-BridgeMT (0.22) & NLLB\_BLEU (0.86)  & \textbf{0.000} \\\midrule

    \end{tabular}%
    }
    \vspace{-5pt}
    \caption*{(b) English $\rightarrow$ German} % Label for the left table
\end{minipage}

\vspace{-8pt}
\caption{Pairwise system rankings and statistical significance of system quality differences for \ZhEn~and \EnDe~under three annotation settings. The results are based on the \textit{z}-normalized scores with the \ZhEn~outlier annotator being excluded. The red highlight points out a system ranking discrepancy. The \textit{p} values  indicate statistical significancy in systems quality, determined by a random permutation test with 10000 trials.}
\label{tab:system_rankings}
\end{table*}

\textbf{MQM, \sxsmqm, and \sxsqr~demonstrate strong agreement in system-level rankings; \sxsqr's high tie rate may impact its reliability.} \tableref{tab:system_rankings} presents the system ranking results. 

For \ZhEn, all three settings yield identical system rankings. For \EnDe, GPT4-5shot was annotated twice in \sxsmqm, once with ONLINE-W and once with refA, and obtained stable scores. This suggests that \sxsmqm~does not compromise the absolute evaluation of individual systems. \sxsqr~on ONLINE-A and ONLINE-Y show a discrepancy with MQM and \sxsmqm; however, the difference is not statistically significant, indicated by the \textit{p}-value.

Further investigation into the discrepancy in \EnDe~\sxsqr~reveals that the high tie rate in \sxsqr~(\tableref{tab:tie_rate}) plays a key role. In three rounds of \sxsqr~annotations, ONLINE-Y and ONLINE-A tied in two. Across all annotations from three annotators, the tie outcome occurred 116 times (37.18\%) in \sxsmqm, compared to 17.31\% in MQM and 29.48\% with \sxsmqm.

% \yscomment{In three rounds of \sxsqr~annotations, ONLINE-Y and ONLINE-A tied in two. Across all annotations from three annotators, the tie outcome occurred 116 times (37.18\%), among which only 12 were shared with MQM and 25 with \sxsmqm.} \prcomment{How about just say ties appeared X\% more than in MQM and Y\% more than in SxS MQM?}

% , compared to 54 and 92 ties in MQM and \sxsmqm, respectively
% /data/yixiao/human_eval_proj/code_sys_ranking/ONLINE-Y_A_ties_per_sys_pair.py
% /data/yixiao/human_eval_proj/code_sys_ranking/ONLINE-Y_A_qr_tie_sxs_not_tie.py
% \prcomment{There are too many numbers embedded in this paragraph. Can we simplify/condense?} %This aligns with the observation in \sectionref{sec:seg_pra} that \sxsqr~tends to have a high tie rate, which offsets its reliability by the lack of explicit error marking, especially for the long and highly similar segments in \EnDe~ONLINE-Y and ONLINE-A.

These findings suggest several insights: (1) the coarse rating scale of \sxsqr~makes it difficult to detect nuanced quality differences; (2) while annotating error in MQM facilitates fine-grained distinctions, it also increases the risk of spurious differences due to rater noise; and (3) \sxsmqm~balances these trade-offs more effectively, as reflected in its higher inter-translation consistency.

\subsection{MQM error distribution}\label{sec:error_distr}

\begin{figure*}[ht]
% https://colab.research.google.com/drive/1oLaOw6CoVY-4Bj_lJlbEAY6CuXkF5Gof?usp=sharing
    \centering
    \includegraphics[scale=0.4]{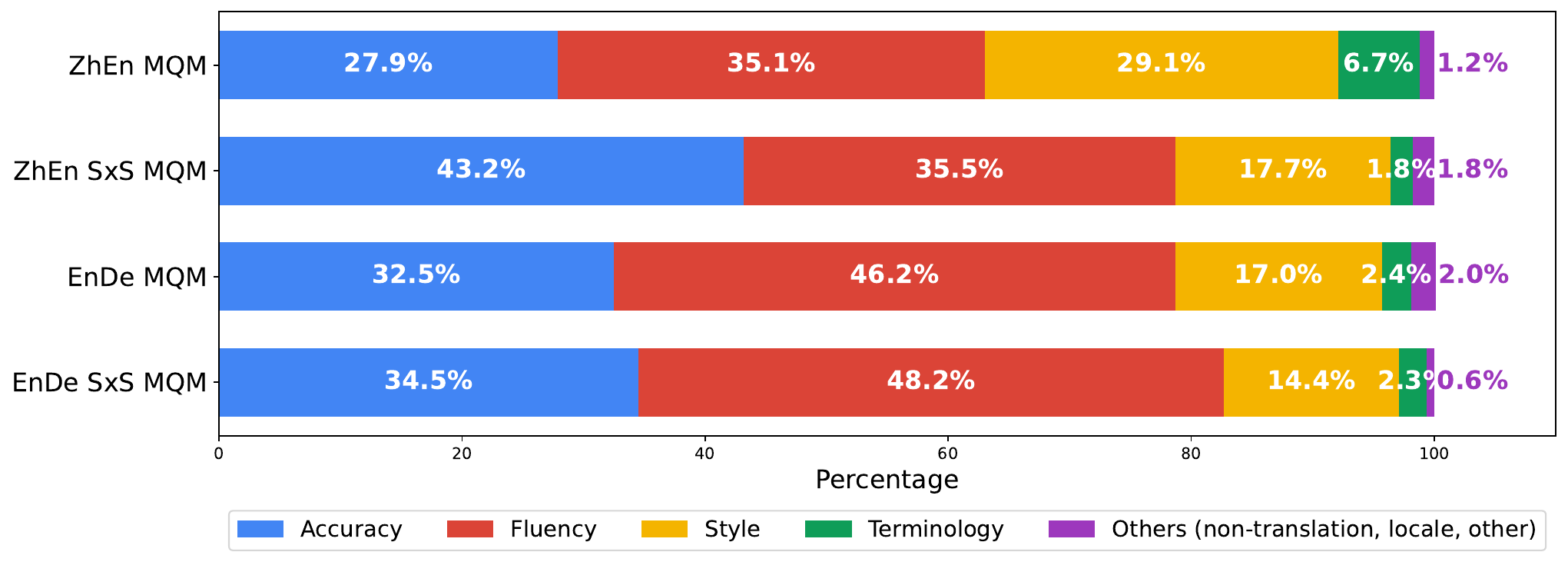}
    \caption{Percentages of error categories in the MQM settings in \ZhEn~and \EnDe. The GPT4-5shot errors in \EnDe~MQM are doubled for a fair comparison with \EnDe~\sxsmqm. While the percentages in \EnDe~stay relatively stable, in \ZhEn, {\color{gglblue} \textbf{accuracy}} errors have a higher percentage in \sxsmqm~than in \psxsmqm.}
    \label{fig:error_perc}
    \vspace{-7pt}
\end{figure*}

\textbf{\sxsmqm~highlights more major accuracy errors in \ZhEn, reflecting its ability in finding accuracy errors that may be neglected in MQM.} \figureref{fig:error_perc} illustrates the distribution of error category percentages for MQM and \sxsmqm.\footnote{Because GPT4-5shot is annotated twice in \EnDe~\sxsmqm, when counting error numbers, the GPT4-5shot errors in MQM are duplicated for a fair comparison between \EnDe~MQM and \sxsmqm.}

\begin{figure*}
% /data/yixiao/human_eval_proj/code_error_distribution/count_error.py
    \centering
    \includegraphics[scale=0.48]{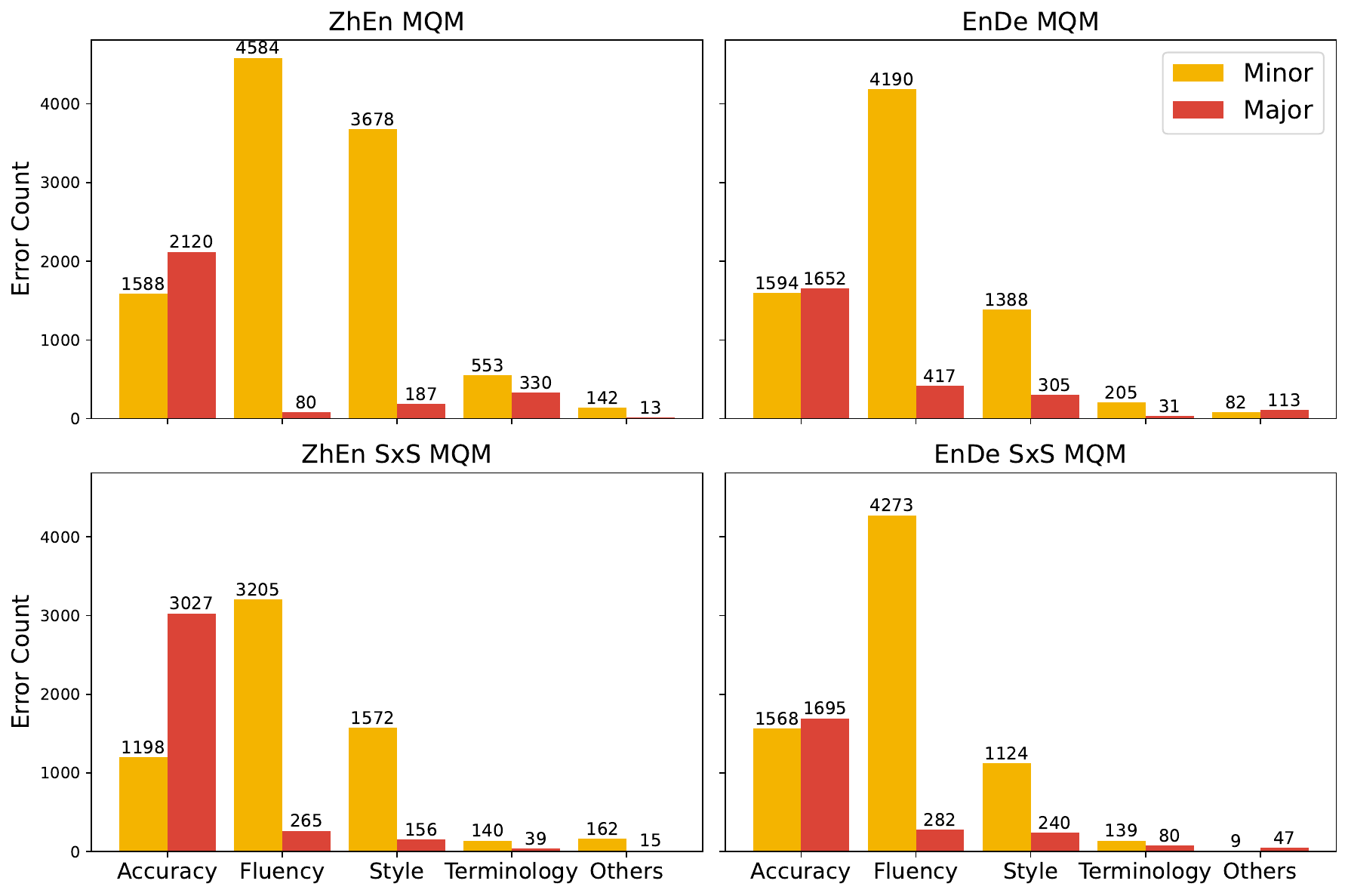}
    \caption{Number of errors in five categories in the MQM settings in \ZhEn~and \EnDe~of all three rounds of annotations. \texttt{Others} includes non-translation, locale convention, and other.
    }
    \label{fig:error_count}
    \vspace{-10pt}
\end{figure*}

While \EnDe~\sxsmqm~shows similar proportions to MQM, \ZhEn~\sxsmqm~shows a higher prevalence of major accuracy errors, further supported by the detailed counts in \figureref{fig:error_count}. To understand the source of major accuracy errors in \ZhEn~\sxsmqm, we examined whether annotators altered their category assignment of the same errors across the two annotation settings. The heatmaps in \figureref{fig:category_conversion} (\appendixref{appendix:error_distribution}) reveal category conversions, notably from Terminology and Style errors in MQM to Accuracy errors in \sxsmqm~for \ZhEn.

The conversions only partially explain the increase in major accuracy errors in \ZhEn. A review of 50 randomly sampled segments from the \ZhEn~ONLINE-A system\footnote{ONLINE-A shows the largest increase in accuracy errors when comparing \sxsmqm~to \psxsmqm.} revealed that many accuracy errors identified in \sxsmqm~were not annotated in \psxsmqm.

Overall, \EnDe~has more fluency errors in both MQM settings while \ZhEn~has a significant increase in accuracy errors in \sxsmqm. This may stem from English and German belonging to the same language family, making fluency the key challenge in translation, whereas the linguistic differences between Chinese and English make accuracy a greater challenge in \ZhEn. \sxsmqm~may further highlight accuracy errors in \ZhEn, especially when only one translation contains such an error.

% \subsection{Suggestions for future human evaluation}

% \yscomment{to be revised}

% Given the results, if comprehensive error identification is the priority, MQM is preferable because \sxsmqm~might inadvertently steer annotators toward focusing on differences that better differentiate systems (e.g., accuracy errors), potentially overlooking minor errors (e.g., fluency). However, \sxsmqm~provides significantly more consistent and reliable error annotation. Lastly, if one's primary goal is system ranking but not detailed error analysis, \sxsqr~offers performance on a par with more sophisticated methods while requiring considerably less time.

% \input{sections/discussion}

\section{Related work}

Human evaluation has long been the gold standard for assessing both MT performance and automatic MT evaluation metrics in the WMT conferences and numerous MT studies. Over the years, it has evolved through efforts to establish more reliable and replicable methods \citep{stanchev-etal-2020-towards}.

\noindent \textbf{Human evaluation in the WMT metrics shared task} transitioned from the 5-point scale on fluency and adequacy \citep{koehn-monz:2006:WMT} to relative ranking (RR) of 5 translation sentences or phrases \citep{callison-burch-etal-2007-meta} for a better inter-/intra-annotator agreement \citep{callison-burch-etal-2007-meta, callison-burch-etal-2008-meta}. WMT16 \citep{bojar-etal-2016-results} used direct assessment (DA) \citep{graham-etal-2013-continuous, graham-etal-2016-glitters, GRAHAM_BALDWIN_MOFFAT_ZOBEL_2017}, where annotators scored translations between 0 and 100. It returns reliable evaluation when each item receives 15 or more judgments. As MT systems advance, DA struggles and mistakenly ranks high-quality human translations below machine outputs \citep{freitag-etal-2021-results}.
% However, DA requires at least 15 judgments per annotation item to achieve reliable evaluation. Furthermore, 
To improve human evaluation quality, MQM \citep{lommel2014mqm, freitag-etal-2021-experts} was introduced into WMT21 \citep{freitag-etal-2021-results}. It emphasizes the inclusion of context \citep{mathur-etal-2020-results} and the use of experts to better capture subtle differences \citep{goto-etal-2014-crowdsourcing, toral-etal-2018-attaining, Lubli2020ASO}. 

\noindent \textbf{MQM is the state-of-the-art MT evaluation method}, but it has shortcomings---disagreement in marking error span boundaries, category, and severity \citep{lommel2014assessing}. By introducing \sxsmqm, the current work aims to address those issues. \citet{kocmi2024esa} aim to mitigate the impact of these issues through Error Span Annotation, a point-wise annotation setting where annotators first identify error spans (with severity) in a segment before assigning it an overall score. The segment-level scores are different from MQM in two ways: first, the scores are assigned to measure the amount of meaning preserved in translations (as in their Figure 1); second, the scores are not automatically calculated from errors, which may introduce subjectivity and latency.

\noindent \textbf{Pairwise evaluation offers a simpler and more intuitive approach to MT evaluation.} \citet{vilar-etal-2007-human} advocate for using binary instead of n-ary RR, as it is more intuitive and straightforward for annotators. Unlike \sxsqr, their method does not provide context and requires annotators to rank segments based on only adequacy and fluency. All possible system pairs are considered, with a full system ranking being obtained either by treating the task as a sorting problem or by applying the Bradley-Terry model \citep{Bradley1952RankAO, dras-2015-squibs}. Pairwise evaluation is also implemented in automatic MT evaluation, for example, by \citet{guzman-etal-2015-pairwise} and \citet{liu2024aligning}.

% draft in progree: \url{https://docs.google.com/document/d/1fqeZsMZnrWgpjGtA6o8l6o8aWZDcHqUFaosdHvZNC00/edit?usp=sharing}

% WMT has been putting efforts in finding the optimal human evaluation methodology.

% \sxsqr~is reminiscent of \citet{vilar-etal-2007-human}: pairwise comparison at the sentence level with no context, full ranking of systems

% DA --> MQM p3. \url{https://wrap.warwick.ac.uk/id/eprint/622/1/WRAP_Stewart_absolute_identification.pdf} judgement of the current stimuli is influenced by the previous one---sequential effects \citep{stewart2005absolute}

% People (experts and non experts) are better at finding the worse/better outputs in a side by side setting \citep{karpinska-etal-2021-perils}

% z-normalization is important if other suggestions are not followed (stability)

% \citet{liu2024aligninghumanjudgementrole}

% \citet{jang2022decreasingannotationburdenpairwise} pairwise in other domains

% \citet{guzman-etal-2015-pairwise}

% \citet{licht-etal-2022-consistent}

% \citet{dras-2015-squibs}

% \citet{freitag-etal-2021-experts}

% \citet{popovic-2021-agree}

% target only MT evaluation \url{https://aclanthology.org/W15-3059.pdf}

% yvette graham 

% ranking five different outputs

% wmt findings paper

% check lit to see if we need to change the name of sxs qr

\section{Conclusion}

This study uses machine translation as a case study and examines the impact of MQM, \sxsmqm, and \sxsqr~on the annotation results from five aspects: inter-annotator agreement, inter-translation error annotation consistency, quality ranking at segment- and system-levels, and error distributions. 

Incorporating comparative judgment, \sxsmqm~and \sxsqr~achieved higher inter-annotator agreement. \sxsmqm~enhanced error marking consistency both for explicitly compared system pairs and across others. Concerning \sxsqr, although it does not provide detailed error annotations, it offers an efficient and reliable alternative for system ranking provided by \sxsmqm, with a trade-off in differentiating subtle quality differences due to higher tie rates.

The findings in the paper demonstrate the value of comparative judgment in improving annotation quality and efficiency, with \sxsmqm~and \sxsqr~serving as practical alternatives to MQM, tailored to different evaluation needs.

\section*{Limitations}

Practical considerations make it difficult to control all variables in human evaluation experiments. As an example, the single-sided MQM annotations were collected in 2023 with a different original goal, while the \sxsmqm~and \sxsqr~annotations were collected in 2024 for this project. Additionally, our annotators were engaged in multiple projects throughout 2024 and thus may have performed other annotation tasks in between items collected for this project.

Due to the time and cost involved in human evaluation, the current work is not able to test MQM, \sxsmqm, and \sxsqr~on a larger set of language pairs besides \ZhEn~and \EnDe. It is also impractical to exhaustively test all possible system pairs in the two tested language pairs. However, the current study still provides a strong foundation for understanding the trade-offs between detailed error detection and overall system ranking. Future work could expand this work to include more language pairs and a broader range of systems, further validating the generalizability of the results.

Future work can also test the annotation setups in other domains beyond MT to see how annotation settings influence evaluations in diverse NLP tasks. Additionally, researchers can investigate the impact of different annotation settings on annotator backgrounds. Although \sxsmqm~and \sxsqr~do not significantly increase the inter-annotator agreement of expert annotators, they might do so for crowd-sourced workers.

\section*{Ethics statement}

Our professional translator annotators were sourced by a translation agency. They were given sufficient time to complete the task and paid fairly. 

The annotators worked on the news data from WMT2023 \citep{freitag-etal-2023-results} that should not contain offensive content. The annotation task did not ask for personally identifiable information.

%NOTE: the \typeout{} command, if put immediately before \bibliography, can sometimes fix problems with all citations being ???, as per https://www.overleaf.com/learn/latex/Questions/BibTeX_isn%27t_working%3B_my_%5Ccite_are_showing_up_as_question_marks_(%3F)
\typeout{}
\bibliography{anthology,custom}
% \bibliography{custom}

\appendix

\section{MQM error categories}\label{appendix:error_hierarchy}

\tableref{tab:error_hierarchy} lists all the error categories and their subcategories mentioned in \sectionref{sec:MQM}. The annotators used the (sub)categories to label errors in the MQM and \sxsmqm~settings.

\begin{table}[ht]
\fontsize{5}{6}\selectfont
\centering
\resizebox{0.9\columnwidth}{!}{%
\begin{tabular}{@{}p{1.2cm}p{2.5cm}@{}}
\midrule
\textbf{Category}   & \textbf{Subcategory}     \\ \midrule
\multirow{6}{*}{\texttt{Accuracy}}  & Reinterpretation \\
                                    & Mistranslation   \\
                                    & Gender Mismatch  \\
                                    & Untranslated     \\
                                    & Addition         \\
                                    & Omission         \\\cmidrule{1-2}
\multirow{7}{*}{\texttt{Fluency}}   & Inconsistency    \\
                                    & Grammar          \\
                                    & Register         \\
                                    & Spelling         \\
                                    & Text-Breaking    \\
                                    & Punctuation      \\
                                    & Character Encoding \\\cmidrule{1-2}
\multirow{3}{*}{\texttt{Style}}     & Unnatural or Awkward    \\
                                    & Bad Sentence Structure  \\
                                    & Archaic or Obscure Word Choice \\\cmidrule{1-2}
\multirow{2}{*}{\texttt{Terminology}}   & Inappropriate for Context  \\
                                        & Inconsistent \\\cmidrule{1-2}
\multirow{6}{*}{\texttt{Locale Convention}}   & Address Format  \\
                                              & Date Format \\
                                              & Currency Format \\
                                              & Telephone Format \\
                                              & Time Format \\
                                              & Name Format \\\cmidrule{1-2}
\multirow{1}{*}{\texttt{Non-Translation}}   & ---  \\\cmidrule{1-2}
\multirow{1}{*}{\texttt{Other}}   & ---  \\\cmidrule{1-2}
\multirow{1}{*}{\texttt{Source Issue}}   & ---  \\\bottomrule

\end{tabular}%
}
\caption{MQM error hierarchy. Error spans are categorized into categories and subcategories. 
}
\label{tab:error_hierarchy}
% \vspace{-10pt}
\end{table}

\section{Annotation scoring scheme}\label{appendix:scoring_scheme}

\tableref{tab:weighting-scheme} 
% and \tableref{tab:weighting-scheme-qr} 
gives the scoring scheme in \sectionref{sec:result_score_calc} of the MQM settings.
% and the \sxsqr~setting respectively.

\begin{table}[ht]
\fontsize{6}{7}\selectfont
\centering
\resizebox{0.7\columnwidth}{!}{%
\begin{tabular}{@{}p{0.55cm}p{1.6cm}p{0.7cm}@{}}
\midrule
\textbf{Severity}   & \textbf{Category} & \textbf{Weight} \\ \midrule
\multirow{2}{*}{\major}  & Non-translation & 25  \\
                         & Others   & 5 \\\cmidrule{1-3}
\multirow{2}{*}{\minor}  & Fluency/Punctuation & 0.1  \\
                         & Others   & 1 \\\midrule
\end{tabular}%
}
\caption{MQM error span weighting scheme. Gibberish segments score 25 points, \major~errors 5 points, and \minor~errors 1 point, except for \minor~punctuation errors, which are weighted at 0.1 point each.
}
\label{tab:weighting-scheme}
\end{table}

% \begin{table}[ht]
% \fontsize{5}{6}\selectfont
% \centering
% % \renewcommand{\arraystretch}{0.8} % Reduce vertical space
% \resizebox{0.6\columnwidth}{!}{%
% \begin{tabular}{@{}p{1.5cm}p{0.7cm}@{}}
% \midrule
% \textbf{Category} & \textbf{Weight} \\ \midrule
% Much worse & 2  \\
% Worse      & 1  \\
% About the same & 0 \\\midrule
% \end{tabular}%
% }
% \caption{\sxsqr~segment scoring scheme.
% }
% \label{tab:weighting-scheme-qr}
% \end{table}

\section{More details to metrics of meta evaluation}
\label{appendix:meta_eval_details}

\subsection{Pairwise ranking agreement}\label{appendix:pra_terms}

\eqref{eq:PRA} is used to calcualate agreement in segment-level rankings between the annotation settings and \tableref{tab:pra_terms} defines its terms. The ranking of two translations of the same source segment depends on the segment scores: segments with identical scores are tied, while differing scores dictate their ranking. PRA quantifies the frequency with which two evaluation settings agree on the ranking of each pair of segments from two systems.

% \vspace{-5pt}
\begin{equation}
  \resizebox{0.6\hsize}{!}{$
  PRA = \frac{C + T_{\alpha\beta}}{C + D + T_\alpha + T_\beta + T_{\alpha\beta}}
  $}
  \label{eq:PRA}
\end{equation}
\vspace{-3pt}

\begin{table}[ht]
% \fontsize{6}{7}\selectfont
\centering
\resizebox{0.85\columnwidth}{!}{%
\begin{tabular}{@{}ll@{}}
\midrule
\textbf{Symbol}    & \textbf{Description}                \\ \midrule
$\alpha$  & One of the annotation settings    \\
$\beta$   & One of the annotation settings that is not $\alpha$ \\
$C$       & The number of concordant pairs    \\
$D$       & The number of discordant pairs    \\
$T_{\alpha}$ & The number of pairs tied \textit{only} in $\alpha$ \\
$T_{\beta}$  & The number of pairs tied \textit{only} in $\beta$ \\
$T_{\alpha\beta}$ & The number of pairs tied in both $\alpha$ and $\beta$ \\\midrule
\end{tabular}%
}
\caption{Terms in \eqref{eq:PRA}. The annotation settings are \psxsmqm, \sxsmqm, and \sxsqr.
}
\label{tab:pra_terms}
\end{table}

\subsection{Inter-translation consistency}\label{appendix:itc_pseudo_code}

Using the example below, inter-translation consistency is meant to be the following: if an annotator labels ``arabica'' as a minor fluency error in \ref{ex:chatgpt}, they should do the same in \ref{ex:claude}.\footnote{The two translations share common spans highlighted in green, identified using the \texttt{get\_opcodes()} function from Python's \texttt{difflib} module.} For reasons of practicality and clarity, two systems are considered at a time for calculating the consistency.

\begin{enumerate}[label=(\alph*)]
    \item {\color{gglgreen}Brazil is the world's largest producer of \underline{\textbf{arabica}} beans,} a {\color{gglgreen}coffee} variety {\color{gglgreen}commonly used by baristas} to make {\color{gglgreen}coffee}.\label{ex:chatgpt}
    \vspace{-6pt}
    
    \item {\color{gglgreen}Brazil is the world's largest producer of \underline{\textbf{arabica}} beans,} which are the {\color{gglgreen}coffee} beans {\color{gglgreen}commonly used by baristas} in making {\color{gglgreen}coffee}.\label{ex:claude}
\end{enumerate}

Inter-translation consistency is calculated as follows: \textbf{Alignment of Tokens} the translations from two systems are tokenized, and the alignment between their tokens is computed using \texttt{get\_opcodes()} from \texttt{difflib}. This generates a list of operations (replace, delete, insert, equal) that align the tokens of the two translations; \textbf{Identification of Potential Common Errors} Errors annotated in each translation are compared based on their spans. If an error in one translation aligns with an ``equal'' operation in the token alignment, it is considered as a potential common error of the two compared translations. These errors are stored for further analysis; \textbf{Matching Errors Using a Criterion} A specified criterion (e.g., matching spans, categories, severities, or combinations thereof) is applied to the potential common errors. This determines how many errors are consistently marked across the two translations; \textbf{Calculation of Consistency} the consistency value is calculated as the percentage of errors that satisfy the criterion out of the total potential errors. 

The final inter-translation consistency is averaged over all raters per criterion.

\section{Identifying and Addressing Outlier Annotators}\label{appendix:outlier_annotator}

One annotator is excluded from the \ZhEn~annotation results due to significant deviations in annotation behavior compared to their peers, which could skew the results and compromise the reliability. This decision was based on two key observations. 

First, this annotator identified an exceptionally high number of errors compared to their peers. Given the mean and the standard deviation of the error counts in \tableref{tab:mean_std_error_cnt}, the outlier annotator's \textit{z}-score for \ZhEn~\sxsmqm~is 2.28, placing them more than two standard deviations above the mean. In contrast, all other annotators, regardless of whether they contributed to \ZhEn~or \EnDe, remain within two standard deviations of the mean.

\begin{table}[ht]
\fontsize{5}{6}\selectfont
\centering
\resizebox{1\columnwidth}{!}{%
\begin{tabular}{rrrr}
\toprule
\textbf{MQM Mean}    & \textbf{MQM Std} & \textbf{\sxsmqm~Mean}    & \textbf{\sxsmqm~Std}  \\\midrule
\multicolumn{4}{c}{\textbf{Chinese $\rightarrow$ English}}\\
2936.1 & 1021.1 & 2642.3 & 1874.1 \\\cmidrule{1-4}

\multicolumn{4}{c}{\textbf{English $\rightarrow$ German}}\\
900.2 & 413.0 & 945.7 & 475.2 \\\bottomrule

\end{tabular}%
}
\caption{
Mean and standard deviation of the error counts from 8 \ZhEn~and 10 \EnDe~annotators.
}
\label{tab:mean_std_error_cnt}
% /data/yixiao/human_eval_proj/code_exclude_56225225/stats.py
\end{table}

\begin{figure*}
    \centering
    \includegraphics[scale=0.51]{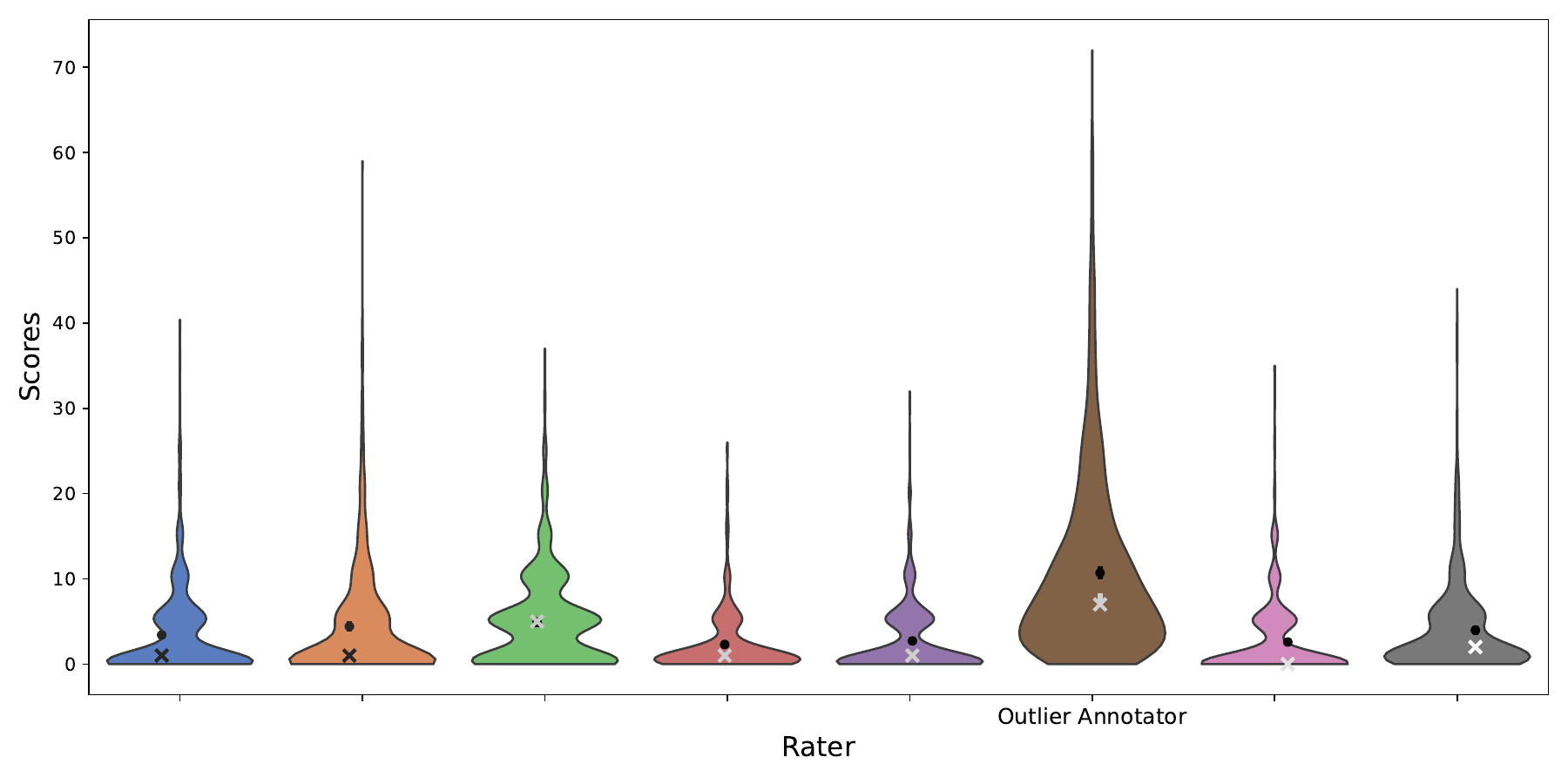}
    \caption{
    Violin plots of the original segment scores contributed by each annotator (without \textit{z}-normalization). Annotator identities are omitted for anonymity. The dots indicate the mean, while the crosses represent the median of each distribution.
    }
    \label{fig:violin}
\end{figure*}

Second, visualizing the segment scores contributed by each annotator using violin plots revealed that the outlier annotator from \ZhEn~was the only one with a distinctly unimodal distribution in \ZhEn~\sxsmqm, as illustrated in \figureref{fig:violin}. This is stark contrast to the more varied distributions observed among other annotators. %This characteristic persisted regardless of whether the scores were \textit{z}-normalized or not, further highlighting a stark deviation from the more varied distributions observed among other annotators. %NOTE FROM PARKER: If I understand correctly, Z-normalization is guaranteed to not affect the multi-/uni-modality of a distribution, so no need to mention it here.

Based on the two key observations of the outlier annotator's behavior in \sxsmqm, we opted to exclude their annotated data from the analyses. After removing the segments annotated by the outlier annotator, there are 16 documents with 220 segments in \ZhEn~with an average token count of 31.45 per segment.

\section{Inter-annotator agreement by pair characteristics}\label{appendix:IAA_expanded}

\tableref{tab:inter-AA-3-buckets} supports the findings made in \sectionref{sec:annotator_agreement} that the annotators have a higher agreement on the shortest segments in the \sxs~setting but the correlation between agreement levels and the segment length is not proportional. The longest segment buckets do not necessarily have the lowest $\alpha$.

\begin{table}[htbp]
    \fontsize{5}{6}\selectfont
    \centering
    \resizebox{0.95\columnwidth}{!}{%
    \begin{tabular}{@{}rrrr@{}}
    \toprule
    \multicolumn{4}{c}{\textbf{Chinese $\rightarrow$ English}} \\
    % Avg. toks (\# seg) & 13.5 (125) & 28.07 (125) & 54.14 (127) \\
    % \textcolor[gray]{0.5}{MQM} & \textcolor[gray]{0.5}{\textbf{0.2265}} & \textcolor[gray]{0.5}{0.1652} & \textcolor[gray]{0.5}{0.2053} \\
    % \sxsmqm & \textbf{0.3178} & 0.2180 & 0.1669 \\
    % \sxsqr  & \textbf{0.3047} & 0.1735 & 0.2093 \\
    Avg. toks (\# seg) & 12.95 (73) & 27.84 (73) & 53.36 (74) \\
    \textcolor[gray]{0.5}{MQM} & \textcolor[gray]{0.5}{\textbf{0.2183}} & \textcolor[gray]{0.5}{0.1739} & \textcolor[gray]{0.5}{0.2021} \\
    \sxsmqm & \textbf{0.3154} & 0.2187 & 0.1708 \\
    \sxsqr  & \textbf{0.2970} & 0.1790 & 0.2084 \\
    \midrule
    
    \multicolumn{4}{c}{\textbf{English $\rightarrow$ German}} \\ 
    Avg. toks (\# seg) & 10.62 (34) & 72.94 (34) & 128.83 (36) \\
    \textcolor[gray]{0.5}{MQM} & \textcolor[gray]{0.5}{0.1930} & \textcolor[gray]{0.5}{0.0956} & \textcolor[gray]{0.5}{\textbf{0.2307}} \\
    \sxsmqm & \textbf{0.3608} & 0.2763 & 0.2849 \\
    \sxsqr  & \textbf{0.3136} & 0.2330 & 0.1645 \\ \bottomrule
    \end{tabular}%
    }
    \caption{Krippendorff's $\alpha$ values for annotators' agreement across three equally sized segment groups of different average token lengths. Shorter segments elicit higher agreement in the \sxs~settings.
    }
    \label{tab:inter-AA-3-buckets}
    % \vspace{-10pt}
\end{table}

\tableref{tab:inter-AA_expanded} expands \tableref{tab:inter-AA} and presents the Krippendorff's $\alpha$ of each type of pairs (i.e., top-two, high textual similarity, and low textual similarity). In both language pairs, when evaluating the top 2 systems, the side-by-side settings achieve higher $\alpha$ than \psxsmqm, with \sxsmqm achieving the highest. The ranking pattern for the 3 settings on the high and low text similarity system pairs is inconsistent between language pairs. For \EnDe, \sxsmqm achieves the highest average $\alpha$ in each group of system pairs.

\begin{table}[htbp]
    \fontsize{5}{6}\selectfont
    \centering
    \resizebox{0.9\columnwidth}{!}{%
    \begin{tabular}{@{}rrrrr@{}}
    \midrule
     & \textbf{All} & \textbf{Top 2} & \textbf{High text sim} & \textbf{Low text sim} \\ \midrule
    \multicolumn{5}{c}{\textbf{Chinese $\rightarrow$ English}} \\
    MQM & 0.2178 & 0.1560 & 0.2118 & 0.2359 \\
    \sxsmqm & 0.2510 & 0.2406 & 0.2290 & 0.2345 \\
    \sxsqr & 0.2380 & 0.2056 & 0.2481 & 0.2336 \\
    
    \multicolumn{5}{c}{\textbf{English $\rightarrow$ Chinese}} \\
    MQM & 0.2345 & 0.1152 & 0.2819 & 0.2213 \\
    \sxsmqm & 0.3594 & 0.2644 & 0.4244 & 0.2947 \\
    \sxsqr & 0.2402 & 0.1604 & 0.2189 & 0.2636 \\\midrule
    \end{tabular}%
    }
    \caption{Krippendorff's $\alpha$ of annotators' agreement in three annotation settings based on all system pairs and system pairs of three characteristics.
    }
    \label{tab:inter-AA_expanded}
    % \vspace{-10pt}
\end{table}

\section{Inter-translation consistency}\label{appendix:ITC}

\tableref{tab:inter-TC_psxs_sxs} presents the inter-translation consistency results for \ZhEn~without removing the outlier annotator. The findings remain consistent with those in \tableref{tab:inter-TC_psxs_sxs}, showing that the outlier annotator also has a higher inter-translation consistency when doing the task using \sxsmqm.

\begin{table*}[ht]
\fontsize{8}{10}\selectfont
\centering
% \renewcommand{\arraystretch}{0.8} % Reduce vertical space
% \resizebox{0.7\columnwidth}{!}{%
% \begin{tabular}{crrrrr}
\begin{tabular}{p{0.5cm}p{1.8cm}p{1.8cm}p{1.8cm}p{1.8cm}p{2.5cm}}
\midrule
 & \textbf{Setting} & \textbf{Span} $\uparrow$ & \textbf{Span + Cat.} $\uparrow$ & \textbf{Span + Sev.} $\uparrow$ & \textbf{Span + Cat. + Sev.} $\uparrow$ \\ \midrule
\multicolumn{6}{c}{Inter-translation consistency from explicitly compared systems (5 pairs)} \\\midrule

\multirow{2}{*}{\ZhEn} & \psxsmqm & 28.65\% & 26.51\% & 27.95\% & 25.99\% \\
 & \cellcolor{green!20}\sxsmqm  & \cellcolor{green!20}68.01\%  & \cellcolor{green!20}67.54\%  & \cellcolor{green!20}67.78\% & \cellcolor{green!20}67.35\% \\\cmidrule{1-6}

\multirow{2}{*}{\EnDe} & \psxsmqm & 45.07\% & 41.61\% & 42.81\% & 40.25\% \\
 & \cellcolor{green!20}\sxsmqm  & \cellcolor{green!20}78.7\%  & \cellcolor{green!20}77.84\% & \cellcolor{green!20}78.48\% & \cellcolor{green!20}77.67\% \\\midrule

\multicolumn{6}{c}{Inter-translation consistency from \underline{\textit{not}} explicitly compared systems (\ZhEn: 40 pairs; \EnDe: 31 pairs)} \\\midrule
\multirow{2}{*}{\ZhEn} & \psxsmqm & 27.84\% & 25.68\% & 27.05\% & 25.11\% \\
 & \cellcolor{green!20}\sxsmqm  & \cellcolor{green!20}47.81\%  & \cellcolor{green!20}46.70\%  & \cellcolor{green!20}46.92\% & \cellcolor{green!20}46.04\% \\\cmidrule{1-6}

\multirow{2}{*}{\EnDe} & \psxsmqm & 45.60\% & 42.87\% & 43.81\% & 41.60\% \\
 & \cellcolor{green!20}\sxsmqm  & \cellcolor{green!20}63.53\%  & \cellcolor{green!20}62.39\% & \cellcolor{green!20}61.67\% & \cellcolor{green!20}60.69\% \\\midrule
\end{tabular}%
% }
\caption{Inter-translation consistency, averaged over 8 (\ZhEn) and 10 (\EnDe) annotators, in \psxsmqm~and \sxsmqm. No annotator's annotation is removed. Cat. = category, Sev. = severity. Inter-translation consistency is calculated for four criteria of what counts as common errors in two systems, for example, Span + Cat. = errors with the same span \textit{and} category. For \EnDe, the annotation of GPT4-5shot in pair with ONLINE-W is not included in the calculation of the lower table results. The green color highlights the higher values between MQM and \sxsmqm.
}
\label{tab:inter-TC_psxs_sxs}
\vspace{-10pt}
\end{table*}

% \begin{table*}[ht]
% \fontsize{8}{10}\selectfont
% \centering
% % \renewcommand{\arraystretch}{0.8} % Reduce vertical space
% % \resizebox{0.7\columnwidth}{!}{%
% \begin{tabular}{crrrrr}
% \midrule
%  & \textbf{Setting} & \textbf{Span} $\uparrow$ & \textbf{Span + Cat.} $\uparrow$ & \textbf{Span + Sev.} $\uparrow$ & \textbf{Span + Cat. + Sev.} $\uparrow$ \\ \midrule
% \multicolumn{6}{c}{Inter-translation consistency from explicitly compared systems (5 pairs)} \\\midrule

% \multirow{2}{*}{\ZhEn} & \psxsmqm & 26.78\% & 24.66\% & 26.18\% & 24.25\% \\
%  & \cellcolor{green!10}\sxsmqm  & \cellcolor{green!10}67.50\%  & \cellcolor{green!10}67.03\%  & \cellcolor{green!10}67.26\% & \cellcolor{green!10}66.85\% \\\midrule

% \multicolumn{6}{c}{Inter-translation consistency from \underline{\textit{not}} explicitly compared systems (\ZhEn: 40 pairs)} \\\midrule
% \multirow{2}{*}{\ZhEn} & \psxsmqm & 26.77\% & 24.63\% & 26.06\% & 24.15\% \\
%  & \cellcolor{green!10}\sxsmqm  & \cellcolor{green!10}46.90\%  & \cellcolor{green!10}45.77\%  & \cellcolor{green!10}46.00\% & \cellcolor{green!10}45.12\% \\\midrule

% \end{tabular}%
% % }
% \caption{The \ZhEn~results of \tableref{tab:inter-TC_psxs_sxs} after removing the outlier annotator, averaged over 7 (\ZhEn).
% }
% \label{tab:inter-TC_psxs_sxs_removed}
% % \vspace{-10pt}
% \end{table*}

\section{Error distribution and category conversion}\label{appendix:error_distribution}

% \figureref{fig:error_count} displays the error counts across five categories (Accuracy, Fluency, Style, Terminology, and Others) for both ZhEn and EnDe MQM and SxS settings. The bars represent the frequency of minor and major errors, illustrating their relative proportions within each category across three rounds of annotations. The 'Others' category includes non-translation, locale convention, and miscellaneous errors. Similar to \figureref{fig:error_perc} (\sectionref{sec:error_distr}), we see a significant increase in \ZhEn~accuracy errors and a modest increase in \EnDe~fluency errors.

% \begin{figure*}
% % /data/yixiao/human_eval_proj/code_error_distribution/count_error.py
%     \centering
%     \includegraphics[scale=0.5]{images/MQM_Error_Count_Comparison_wo_56225225.pdf}
%     \caption{Number of errors in five categories in the MQM settings in \ZhEn~and \EnDe~of all three rounds of annotations. \texttt{Others} includes non-translation, locale convention, and other.
%     }
%     \label{fig:error_count}
% \end{figure*}

\begin{figure*}[ht]
% /data/yixiao/human_eval_proj/code_error_distribution/error_cat_change.py
% old plot: https://colab.research.google.com/drive/1oLaOw6CoVY-4Bj_lJlbEAY6CuXkF5Gof?usp=sharing
    \centering
    % Subfigure (a) ZhEn Heatmap
    \begin{subfigure}[b]{0.45\textwidth}
        \centering
        \includegraphics[width=\textwidth]{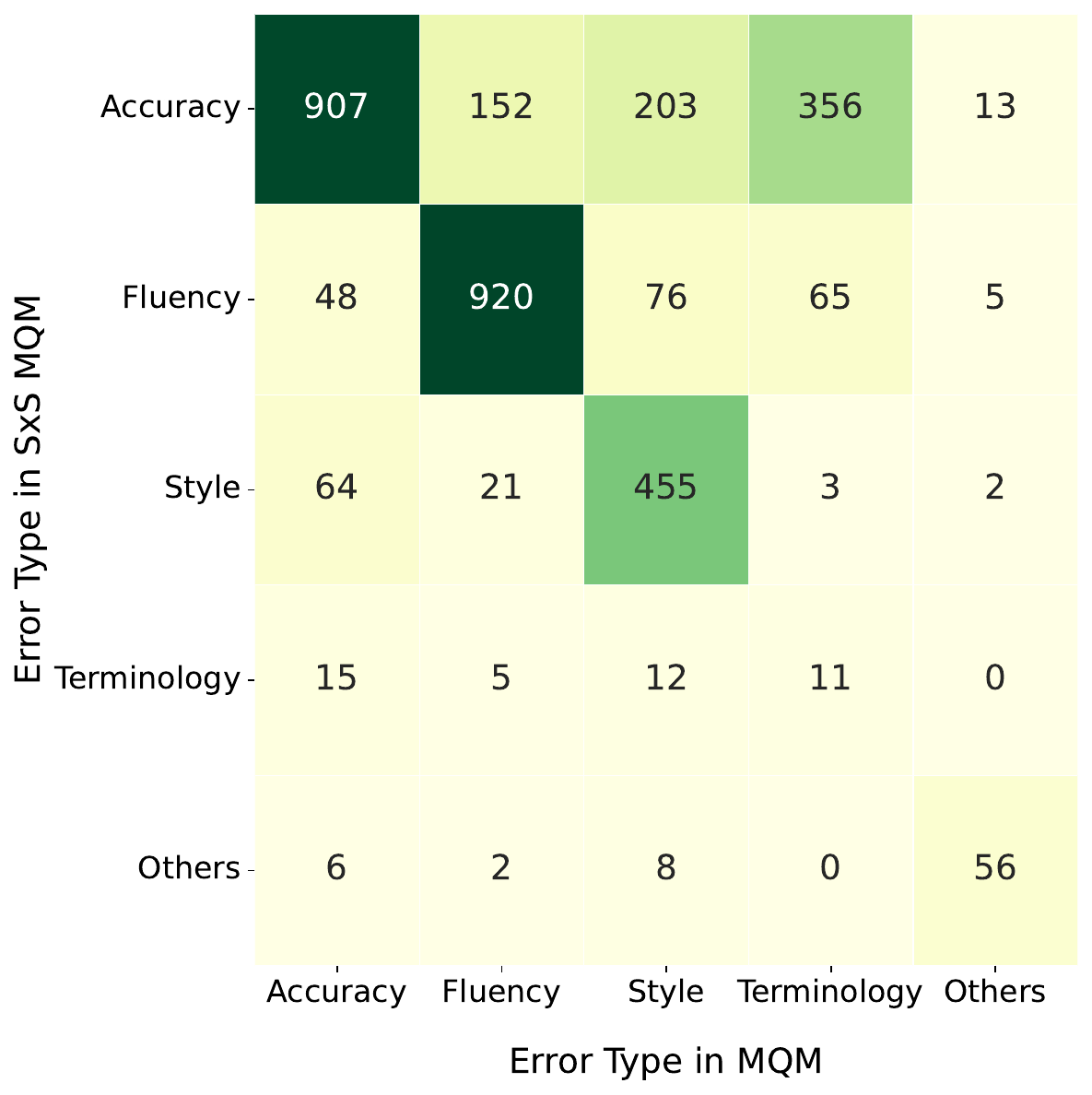}
        \caption{ZhEn error category conversion}
        \label{fig:ZhEn_heatmap}
    \end{subfigure}
    \hspace{0.05\textwidth}  % Horizontal space between figures
    % Subfigure (b) EnDe Heatmap
    \begin{subfigure}[b]{0.46\textwidth}
        \centering
        \includegraphics[width=\textwidth]{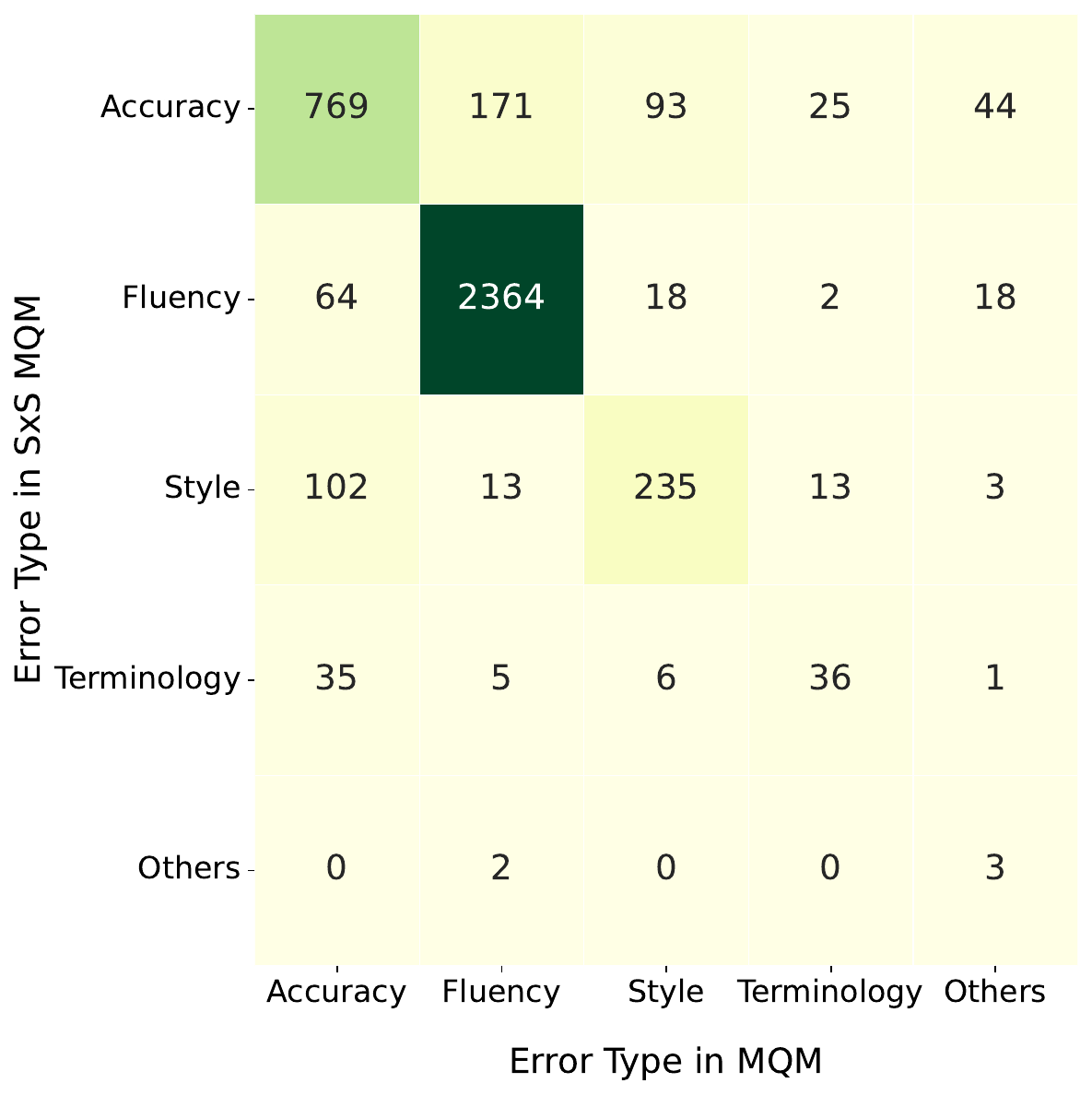}
        \caption{EnDe error category conversion}
        \label{fig:EnDe_heatmap}
    \end{subfigure}
    
    \caption{Error category conversion from \psxsmqm~to \sxsmqm~in (a) \ZhEn~and (b) \EnDe~of the same errors annotated by the annotators in both MQM and \sxsmqm. MQM \EnDe~GPT4-5shot is duplicated for the comparison.}
    \label{fig:category_conversion}
\end{figure*}

The heatmaps in \figureref{fig:category_conversion} demonstrate that there are some error category conversion between MQM and \sxsmqm, which is more prominent in \ZhEn~than in \EnDe. 

\end{document}